\documentclass[11pt]{article}
\usepackage[utf8]{inputenc}
\usepackage[final]{acl}
\usepackage{multirow, makecell}
\usepackage{amsmath}
\usepackage{amssymb}
\usepackage{enumitem}

\usepackage{times}
\usepackage{latexsym}
\usepackage{graphicx}

\usepackage{xcolor,colortbl}
\usepackage[T1]{fontenc}
\usepackage{booktabs}
\usepackage{tikz}
\usetikzlibrary{shapes.geometric, arrows.meta, positioning, fit, backgrounds}
\usepackage{microtype}
\usepackage{subcaption}
\usepackage[most]{tcolorbox}

\title{NCL-UoR at SemEval-2026 Task 5: Embedding-Based Methods, Fine-Tuning, and LLMs for Word Sense Plausibility Rating}
\author{
  Tong Wu\textsuperscript{1}, Thanet Markchom\textsuperscript{2}, and Huizhi Liang\textsuperscript{3} \\
  \textsuperscript{1}Independent Researcher \\
  \textsuperscript{2}Department of Computer Science, University of Reading, Reading, UK \\
  \textsuperscript{3}School of Computing, Newcastle University, Newcastle upon Tyne, UK \\
  \texttt{tongwuwhitney@gmail.com}, 
  \texttt{thanet.markchom@reading.ac.uk}, \\
  \texttt{huizhi.liang@newcastle.ac.uk}
}

\begin{document}
\maketitle

\begin{abstract}
Word sense plausibility rating requires predicting the human-perceived plausibility of a given word sense on a 1--5 scale in the context of short narrative stories containing ambiguous homonyms. This paper systematically compares three approaches: (1)~\textbf{embedding-based methods} pairing sentence embeddings with standard regressors, (2)~\textbf{transformer fine-tuning} with parameter-efficient adaptation, and (3)~\textbf{large language model (LLM) prompting} with structured reasoning and explicit decision rules. The best-performing system employs a structured prompting strategy that decomposes evaluation into narrative components (precontext, target sentence, ending) and applies explicit decision rules for rating calibration. The analysis reveals that structured prompting with decision rules outperforms both fine-tuned models and embedding-based approaches, and that prompt design matters more than model scale for this task.
\end{abstract}

\section{Introduction}
Word Sense Disambiguation (WSD) has traditionally been framed as selecting the single correct sense for a word in context \cite{navigli2009word}. Yet real-world language is often ambiguous, and multiple senses may be plausible with varying degrees of contextual support \cite{erk2009graded}. SemEval-2026 Task~5 \cite{semeval2026-task-5} addresses this gap through the AmbiStory dataset \cite{gehring-roth-2025-ambistory}, which reframes WSD as a \textit{graded plausibility rating} task over English narratives. Given a five-sentence story with an ambiguous homonym, systems must predict the human-perceived plausibility of a specific word sense on a 1--5 scale.

Three distinct modeling approaches are investigated: (1)~\textit{embedding-based methods} extracting similarity features from sentence embeddings for use with classical regressors; (2)~\textit{transformer fine-tuning} adapting pre-trained language models with LoRA for plausibility regression; and (3)~\textit{LLM prompting} using structured reasoning prompts with explicit evaluation criteria and calibration rules. The core strategy decomposes plausibility judgment into component-level evaluations of precontext, target sentence, and ending, then combines them into a final rating via decision rules.

On the test set, GPT-4o with structured prompting achieves $\rho = 0.731$ and Acc.\ = 0.794, outperforming fine-tuned and embedding-based methods. The code is available\footnote{\url{https://github.com/tongwu17/SemEval-2026-Task5}}.

\section{Background}
\paragraph{Task Description.}
Given a five-sentence English narrative containing an ambiguous homonym, SemEval-2026 Task~5 \cite{semeval2026-task-5} asks systems to predict the human-perceived plausibility of a specific word sense on a 1--5 scale. The task uses the AmbiStory dataset \cite{gehring-roth-2025-ambistory}, a collection of short narrative stories designed to probe lexical ambiguity. Each story consists of a three-sentence precontext that establishes the narrative setting, a target sentence containing the homonym, and an ending sentence that may disambiguate it toward one of its senses. The input is the full narrative with a candidate word meaning; the output is the plausibility rating.

As a concrete example, consider the homonym \textit{ring} with the candidate meaning ``a characteristic sound.''  The story reads: \textit{John looked at his savings and smiled. He had been careful with his money for months. Now, he finally felt ready to make a big decision for their anniversary. He told his girlfriend he would give her a ring. John was excited to finally buy the special piece of jewelry.} The ending confirms the \textit{jewelry} sense, so ``a characteristic sound'' is implausible (gold~=~1).

Each sample was rated by at least five annotators; the gold label is the average rating. The dataset contains 2,280 training, 588 development, and 930 test samples (all in English). 

\paragraph{Word Sense Disambiguation.}
The graded plausibility formulation departs from traditional WSD, which assumes a single correct sense per context \cite{navigli2009word}. Graded word sense assignment \cite{erk2009graded} acknowledges that senses exist on a plausibility continuum, motivating the regression formulation adopted in this work. This naturally motivates casting the task as graded plausibility regression.
 
\paragraph{Transformer Fine-Tuning.}
ELECTRA \cite{clark2020electra} introduces replaced token detection, a sample-efficient pre-training task that trains a discriminator over all input tokens rather than a masked subset. DeBERTa \cite{he2021deberta} improves attention with disentangled content and position representations. LoRA \cite{hu2022lora} enables parameter-efficient adaptation by injecting low-rank decomposition matrices into frozen pre-trained weights.

\paragraph{Sentence Embeddings.}

For the embedding-based paradigm, Sentence-BERT \cite{reimers2019sentencebert} derives semantically meaningful sentence embeddings via siamese networks, enabling efficient similarity computation between story contexts and word sense descriptions without task-specific fine-tuning. The resulting vectors serve as input features for downstream regressors.

\paragraph{LLM Prompting.}

Scaling language models enables few-shot task performance without fine-tuning \cite{brown2020language}. Prompt design and structured instructions play a central role in task performance \cite{liu2023pretrain}. Recent work on LLM-as-a-judge \cite{zheng2023judging} shows that LLMs with structured evaluation criteria can approximate human judgments. This work uses several OpenAI GPT models (GPT-4o\footnote{gpt-4o-2024-08-06}, GPT-4.1\footnote{gpt-4.1-2025-04-14}, GPT-5 mini\footnote{gpt-5-mini-2025-08-07}, GPT-5.2\footnote{gpt-5.2-2025-12-11}), Llama~3.2\footnote{llama-3.2-3B-Instruct} \cite{grattafiori2024llama3}, and Ministral\footnote{ministral-3-8B-Instruct-2512}\cite{mistral2026ministral}, exploring structured prompting strategies with explicit evaluation criteria and decision rules for graded plausibility prediction.

\section{Methodology}

Three approaches are investigated for predicting word sense plausibility ratings. Each takes as input a narrative story, a homonym, and a candidate word sense, and outputs a plausibility rating from 1 to 5. Figure~\ref{fig:system} illustrates the overall system architecture.

\begin{figure*}[t]
\centering
\begin{tikzpicture}[
  node distance=0.4cm and 0.4cm,
  every node/.style={font=\small},
  box/.style={draw, rounded corners, minimum height=0.65cm, minimum width=1.5cm, align=center},
  input/.style={box, fill=blue!10},
  process/.style={box, fill=orange!15},
  output/.style={box, fill=green!15},
  arrow/.style={-{Stealth[length=2mm]}, thick},
]

\node[font=\small\bfseries, anchor=east] (lab1) at (0, 0) {(a) Embedding-Based};
\node[input, right=0.5cm of lab1] (in1) {Story + Sense};
\node[process, right=of in1] (emb) {Sentence\\Embeddings};
\node[process, right=of emb] (feat) {Feature\\Extraction};
\node[process, right=of feat] (reg) {Ridge / XGBoost};
\node[output, right=of reg] (out1) {Rating};

\draw[arrow] (in1) -- (emb);
\draw[arrow] (emb) -- (feat);
\draw[arrow] (feat) -- (reg);
\draw[arrow] (reg) -- (out1);

\node[font=\small\bfseries, anchor=east] (lab2) at (0, -1.5) {(b) Fine-Tuning};
\node[input, right=0.5cm of lab2] (in2) {Story + Sense};
\node[process, right=of in2] (enc) {Transformer\\Encoder};
\node[process, right=of enc] (lora) {LoRA\\Adapter};
\node[process, right=of lora] (head) {Regression\\Head};
\node[output, right=of head] (out2) {Rating};

\draw[arrow] (in2) -- (enc);
\draw[arrow] (enc) -- (lora);
\draw[arrow] (lora) -- (head);
\draw[arrow] (head) -- (out2);

\node[font=\small\bfseries, anchor=east] (lab3) at (0, -3.0) {(c) LLM Prompting};
\node[input, right=0.5cm of lab3] (in3) {Story + Sense};
\node[process, right=of in3] (prompt) {Structured\\Prompt};
\node[process, right=of prompt] (llm) {LLM};
\node[output, right=of llm] (out3) {Rating};

\draw[arrow] (in3) -- (prompt);
\draw[arrow] (prompt) -- (llm);
\draw[arrow] (llm) -- (out3);

\end{tikzpicture}
\caption{System overview of the three approaches. All approaches take the same input (narrative story, homonym, candidate word sense) and output a plausibility rating from 1 to 5.}
\label{fig:system}
\end{figure*}

\subsection{Embedding-Based Methods}
\label{sec:embedding}

\paragraph{MPNet + Ridge Regression.}
The story and candidate meaning are encoded into sentence embeddings using \texttt{all-mpnet-base-v2} \cite{song2020mpnet, reimers2019sentencebert}. From these, 8 features are extracted: cosine similarity, Euclidean distance, and dot product between embeddings, text length features, a binary ending indicator, and interaction terms. These are fed into a Ridge regressor \cite{hoerl1970ridge} with $\alpha = 1.0$.

\paragraph{RoBERTa + XGBoost.}

As an alternative configuration, 23 features are extracted using \texttt{all-roberta-large-v1}, a RoBERTa-based \cite{liu2019robertarobustlyoptimizedbert} sentence embedding model \cite{reimers2019sentencebert}: similarity features (cosine, Euclidean, Manhattan, dot product), lexical overlap (word overlap, Jaccard, character overlap), structural features (sentence/punctuation counts), and interaction terms. An XGBRegressor \cite{chen2016xgboost} is used with regularization and Spearman-based early stopping.

\subsection{Transformer Fine-Tuning}
\label{sec:finetuning}

Two families of models are fine-tuned with LoRA \cite{hu2022lora} for regression.

\paragraph{ELECTRA-base and ELECTRA-large + LoRA.}
Two ELECTRA variants \cite{clark2020electra} are fine-tuned: ELECTRA-base (110M parameters, full fine-tuning) and ELECTRA-large (335M parameters, LoRA with $r{=}8$, $\alpha{=}32$). The input format is \texttt{[meaning] [SEP] [story]} with labels normalized from 1--5 to 0--1. For ELECTRA-large, mean pooling over all tokens (instead of only \texttt{[CLS]}) and Huber loss ($\delta{=}1.0$) \cite{huber1964robust} are applied for robustness to annotator disagreement. Training uses batch size 32 with early stopping (patience 3) based on Spearman correlation.

\paragraph{DeBERTa-large + LoRA with Pairwise and Uncertainty Losses.}
DeBERTa-large \cite{he2021deberta} is fine-tuned with LoRA. The input concatenates the precontext, target sentence, and ending, separated by \texttt{[SEP]} from the word sense description, with three pooling methods considered: \texttt{[CLS]} token pooling, mean pooling, and attention-based pooling. Beyond standard regression loss, two additional loss components are introduced to better optimize for the evaluation metrics:

\begin{itemize}[nosep,leftmargin=*]
    \item \textbf{RankNet pairwise loss} \cite{burges2005learning}: Since Spearman correlation measures rank correlation, a pairwise ranking loss is added that encourages the model to correctly order sample pairs by plausibility. For a mini-batch, pairs $(i, j)$ where $y_i > y_j$ are sampled, and the loss $\mathcal{L}_{\text{rank}} = -\log\sigma(\hat{y}_i - \hat{y}_j)$ is optimized.

     \item \textbf{Uncertainty-aware loss}: To incorporate human uncertainty in word sense plausibility ratings, the annotator standard deviation is used as a tolerance margin during training. Prediction errors within the human disagreement range incur no penalty, while errors exceeding this range are penalized linearly. Specifically, for each sample $i$, $\mathcal{L}_{\text{unc}} = \max\left(0,\; \left|\hat{y}_i - y_i\right| - \sigma_i\right)$, where $\sigma_i$ denotes the standard deviation of annotators' scores for that instance.
\end{itemize}

The total loss is $\mathcal{L} = \mathcal{L}_{\text{reg}} + \lambda_r \mathcal{L}_{\text{rank}} + \lambda_u \mathcal{L}_{\text{unc}}$, where $\lambda_r$ and $\lambda_u$ are weighting hyperparameters.

\subsection{LLM Prompting}
\label{sec:prompting}

Two prompting strategies are designed for LLMs.

\paragraph{Few-Shot Prompting (P1).}
The prompt consists of (1) a system message defining the task and 1--5 rating scale, emphasizing that ``the ending is the most important factor for disambiguation''; (2) five few-shot examples selected from training data (one per rating level, choosing samples with zero annotator standard deviation); and (3) the user prompt with the target sample. Temperature 0 is used for deterministic output.

\paragraph{Structured Prompting with Decision Rules (P2).}
\label{sec:P2}
An improved prompt is designed that replaces few-shot examples with structured evaluation criteria and explicit decision rules:

\begin{enumerate}[nosep,leftmargin=*]
    \item \textit{Component-wise evaluation}: The prompt instructs the model to separately evaluate three narrative components: precontext (``does the setup make this meaning likely?''), target sentence (``does the local usage support this meaning?''), and ending (``does the conclusion reinforce this meaning?''). The ending is identified as ``the strongest source of evidence.''
    \item \textit{Decision rules}: Explicit calibration rules constrain the rating: (a) ``if the ending clearly contradicts the proposed meaning, the rating must be 1 or 2''; (b) ``if evidence is mixed or unclear, choose the lower plausible rating''; (c) ``a rating of 5 requires explicit confirmation in the ending and no contradictions elsewhere.''
    \item \textit{Impartial framing}: The prompt positions the model as ``an impartial evaluator'' who should ``base judgment only on the text given,'' reducing potential biases.
\end{enumerate}

This strategy produces more calibrated predictions without requiring examples in the prompt.

\section{Experimental Setup} 

\paragraph{Data Usage.}
Models are trained on the training set for development experiments. For test evaluation, fine-tuned and embedding-based models are retrained on combined train+dev data with an internal validation split (10--20\%) for early stopping.

\paragraph{Evaluation Metrics.}
Two metrics are reported: (1) \textbf{Spearman correlation} ($\rho$) between predicted and gold ratings, measuring rank-order agreement; and (2) \textbf{accuracy}, defined as the proportion of predictions falling within one standard deviation of the mean annotator rating.

\paragraph{Implementation.}
Fine-tuning experiments used HuggingFace Transformers \cite{wolf2020transformers} with LoRA via PEFT. LLM predictions were obtained via the OpenAI API for GPT models and locally via HuggingFace Transformers for open-source models (Llama, Ministral). 

For the DeBERTa-large + LoRA system, the regression objective was evaluated using both Mean Squared Error (MSE) and Huber loss. The ranking loss weight $\lambda_r$ was set to 0.25 or 0.5, while the uncertainty-aware loss weight $\lambda_u$ was varied among 0.1, 0.3, and 0.5.
For the LoRA component, the rank $r$ was set to 4, 8, or 12, with $\alpha = 32$ and a dropout rate of 0.1. All other training parameters were held constant, including a batch size of 8, 10 training epochs, a learning rate of 1e-4, a warmup ratio of 0.1, and a weight decay of 0.01.
Model selection was performed based on the highest average of Spearman correlation and accuracy on the development set, without cross-validation. The final model used mean pooling, MSE loss, $\lambda_r = 0.25$, $\lambda_u = 0.5$, and $r = 8$.

\section{Results}

\subsection{Development Set Results}

Table~\ref{tab:dev-results} presents results on the development set.

\begin{table}[t]
\centering
\small
\setlength{\tabcolsep}{5pt}
\begin{tabular}{@{}llcc@{}}
\toprule
\textbf{Approach} & \textbf{System} & $\boldsymbol{\rho}$ & \textbf{Acc.} \\
\midrule
\multirow{2}{*}{Embedding} 
  & MPNet + Ridge & 0.174 & 0.560 \\
  & RoBERTa + XGBoost & 0.114 & 0.537 \\
\midrule
\multirow{4}{*}{Fine-tuning}
  & ELECTRA-base & 0.491 & 0.663 \\
  & ELECTRA-large + LoRA & 0.644 & 0.709 \\
  & DeBERTa-large + LoRA & 0.587 & 0.757 \\
  & \quad + uncertainty loss & 0.606 & 0.767 \\
\midrule
\multirow{7}{*}{Prompting}
  & Llama-3.2-3B (P2) & 0.108 & 0.526 \\
  & Ministral-3-8B (P2) & 0.441 & 0.628 \\
  & GPT-5.2 (P1) & 0.634 & 0.721 \\
  & GPT-5 mini (P2) & 0.716 & 0.760 \\
  & GPT-5.2 (P2) & 0.727 & 0.781 \\
  & GPT-4.1 (P2) & 0.726 & 0.769 \\
  & GPT-4o (P2) & \textbf{0.749} & \textbf{0.818} \\
\bottomrule\end{tabular}
\caption{Development set results. P1 = few-shot prompting; P2 = structured prompting with decision rules.}
\label{tab:dev-results}
\end{table}

\subsection{Test Set Results}
Table~\ref{tab:test-results} shows test set results. The best system (GPT-4o with P2) ranked 9th on the leaderboard.

\begin{table}[t]
\centering
\small
\setlength{\tabcolsep}{5pt}
\begin{tabular}{@{}llcc@{}}
\toprule
\textbf{Approach} & \textbf{System} & $\boldsymbol{\rho}$ & \textbf{Acc.} \\
\midrule
\multirow{2}{*}{Embedding}
  & MPNet + Ridge & 0.109 & 0.513 \\
  & RoBERTa + XGBoost & 0.110 & 0.505 \\
\midrule
\multirow{4}{*}{Fine-tuning}
  & ELECTRA-base & 0.482 & 0.625 \\
  & ELECTRA-large + LoRA & 0.527 & 0.639 \\
  & DeBERTa-large + LoRA & 0.492 & 0.676 \\
  & \quad + uncertainty loss & 0.435 & 0.659 \\
\midrule
\multirow{7}{*}{Prompting}
  & Llama-3.2-3B (P2) & 0.134 & 0.522 \\
  & Ministral-3-8B (P2) & 0.472 & 0.594 \\
  & GPT-5.2 (P1) & 0.635 & 0.713 \\
  & GPT-5 mini (P2) & 0.696 & 0.743 \\
  & GPT-5.2 (P2) & 0.717 & 0.760 \\
  & GPT-4.1 (P2) & 0.722 & 0.767 \\
  & GPT-4o (P2) & \textbf{0.731} & \textbf{0.794} \\
\bottomrule
\end{tabular}
\caption{Test set results. P1 = few-shot prompting; P2 = structured prompting with decision rules.}
\label{tab:test-results}
\end{table}

\begin{figure}[t]
\centering
\includegraphics[width=\linewidth]{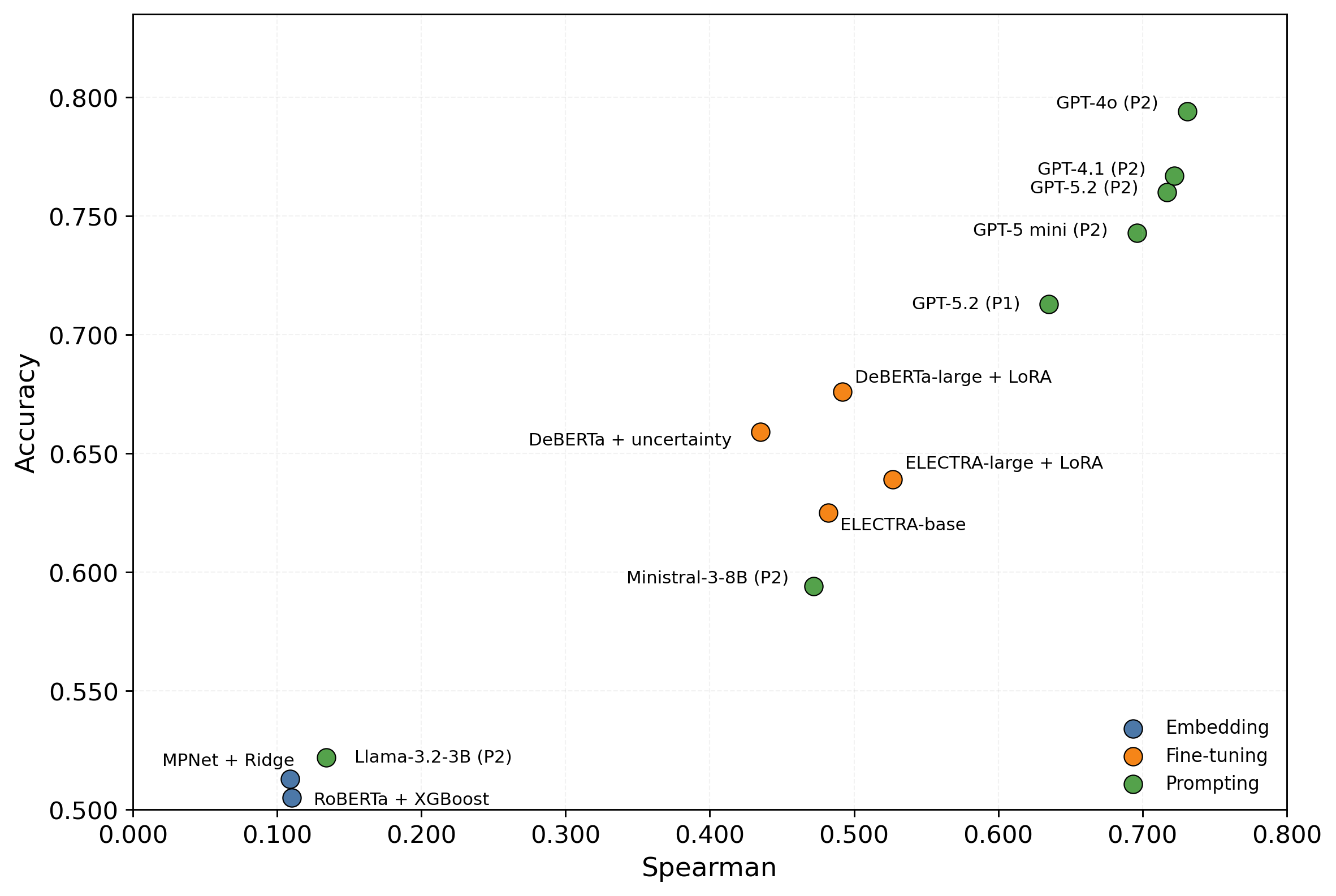}
\caption{Test-set performance landscape (Spearman vs. Accuracy).}
\label{fig:test-tradeoff}
\end{figure}

\paragraph{Embedding features are insufficient for narrative reasoning.}
Both embedding-based methods achieve very low Spearman correlations ($\rho < 0.18$ on dev, $< 0.14$ on test), despite using rich feature sets. This suggests that handcrafted similarity features between story embeddings and meaning embeddings cannot capture \textit{how} a narrative context supports or contradicts a specific word interpretation. The task requires compositional reasoning across multiple sentences, which static similarity metrics fail to model.

\paragraph{Fine-tuning captures contextual dependencies.}
Fine-tuned models substantially outperform embedding methods. ELECTRA-large + LoRA ($\rho = 0.644$ on dev) benefits from larger model capacity and mean pooling over all tokens. DeBERTa-large + LoRA with uncertainty loss achieves the best fine-tuning accuracy (0.767 on dev) by learning to down-weight high-disagreement samples. However, fine-tuning performance degrades on the test set ($\rho = 0.527$ for ELECTRA-large), suggesting difficulty in generalizing to unseen homonyms and story patterns. The substantial dev-to-test drop for ELECTRA-large + LoRA, from $\rho = 0.644$ to $0.527$, likely reflects overfitting to the specific homonyms present in the development set: the model was retrained on combined train+dev data with only a small validation split (10--20\%), limiting its ability to generalize plausibility representations to novel lexical ambiguities in the test set.

\paragraph{Structured prompting outperforms few-shot prompting.}
For GPT-5.2, switching from few-shot prompting (P1: $\rho = 0.635$) to structured prompting with decision rules (P2: $\rho = 0.717$) yields a 0.082 improvement on the test set. The decision rules provide explicit calibration (e.g., ``if the ending contradicts, rate 1--2''), which aligns model predictions with the annotation guidelines. This replaces the need for example memorization with principled reasoning.

\paragraph{Prompt design matters more than model scale.}
On the test set, GPT-4o with structured prompting ($\rho = 0.731$) outperforms GPT-5.2 with the same prompt ($\rho = 0.717$), suggesting that for this task, GPT-4o's reasoning capabilities are well-matched to the structured evaluation framework. 

\begin{figure}[t]
\centering
\includegraphics[width=\linewidth]{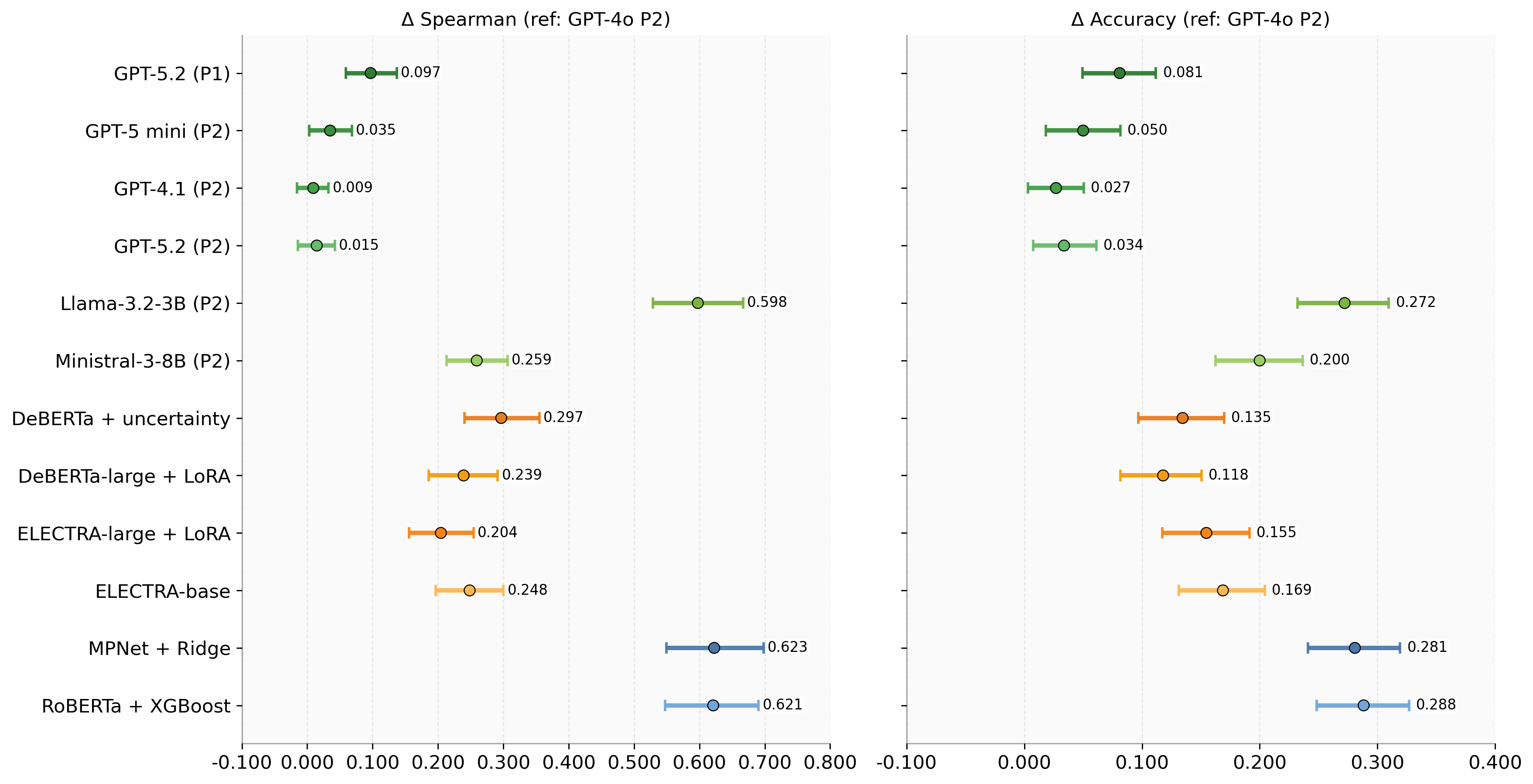}
\caption{Paired bootstrap differences relative to GPT-4o (P2) with bootstrap confidence intervals.}
\label{fig:significance-forest}
\end{figure}

\paragraph{Statistical significance testing.}
Figure~\ref{fig:significance-forest} shows paired bootstrap point estimates and 95\% CIs relative to GPT-4o (P2). The tests ($B{=}2000$), under a paired resampling protocol with fixed random seed, show that GPT-4o is significantly better than GPT-5.2 (P1) on both Spearman ($\Delta\rho = 0.097$, 95\% CI $[0.059, 0.137]$, $p < 0.001$) and accuracy ($\Delta\mathrm{Acc.}=0.081$, $p < 0.001$). For GPT-4.1 (P2) and GPT-5.2 (P2), Spearman differences from GPT-4o are not significant ($p=0.445$ and $p=0.317$), though accuracy gains are small but significant ($p<0.05$). GPT-5 mini (P2) is also significantly worse on both metrics ($p<0.05$), as are both DeBERTa variants ($p{<}0.001$). Llama-3.2-3B (P2), Ministral-3-8B (P2), ELECTRA-large + LoRA, ELECTRA-base, MPNet + Ridge, and RoBERTa + XGBoost are worse on both metrics ($p < 0.001$). Overall, these results indicate that among GPT variants, prompt design rather than model scale is the main driver, and that the LLM prompting advantage over fine-tuning and embedding approaches is robust rather than due to chance.

\subsection{Ablation: Loss Components for DeBERTa}

To better understand the contribution of each auxiliary objective to the fine-tuning behavior, Table~\ref{tab:ablation} ablates the loss components for DeBERTa-large + LoRA on the development set.

\begin{table}[t]
\centering
\small
\begin{tabular}{@{}lcc@{}}
\toprule
\textbf{Loss Configuration} & $\boldsymbol{\rho}$ & \textbf{Acc.} \\
\midrule

Huber + RankNet ($\lambda_r{=}0.5$) & 0.587 & 0.757 \\
\makecell[l]{MSE + RankNet + Uncertainty \\ \quad ($\lambda_r{=}0.25$, $\lambda_u{=}0.5$)} & 0.606 & 0.767 \\
\bottomrule
\end{tabular}
\caption{Ablation of loss components for DeBERTa-large + LoRA on the development set.}
\label{tab:ablation}
\end{table}

Adding uncertainty loss improves both metrics on the development set, confirming that modeling annotator disagreement through learned uncertainty is beneficial. However, uncertainty loss degrades test set performance: DeBERTa-large + LoRA drops from $\rho = 0.492$ without uncertainty loss to $\rho = 0.435$ with it. The similar standard deviation distributions between development and test sets suggest this drop is unlikely to come from distribution shift and more likely reflects mild overfitting during development-set model selection, reducing generalization to unseen test examples.

\subsection{Error Analysis}
\paragraph{High annotator disagreement increases prediction difficulty.}
On the test set, samples with high annotator standard deviation ($\sigma \geq 1.0$; $n = 420$) have a mean absolute error (MAE) of 0.962, compared to 0.765 for low-disagreement samples ($\sigma < 1.0$; $n = 510$). 

\begin{samepage}
\begin{center}
\begin{minipage}{\linewidth}
\centering
\begin{minipage}[t]{0.49\linewidth}
  \centering
  \includegraphics[width=\linewidth]{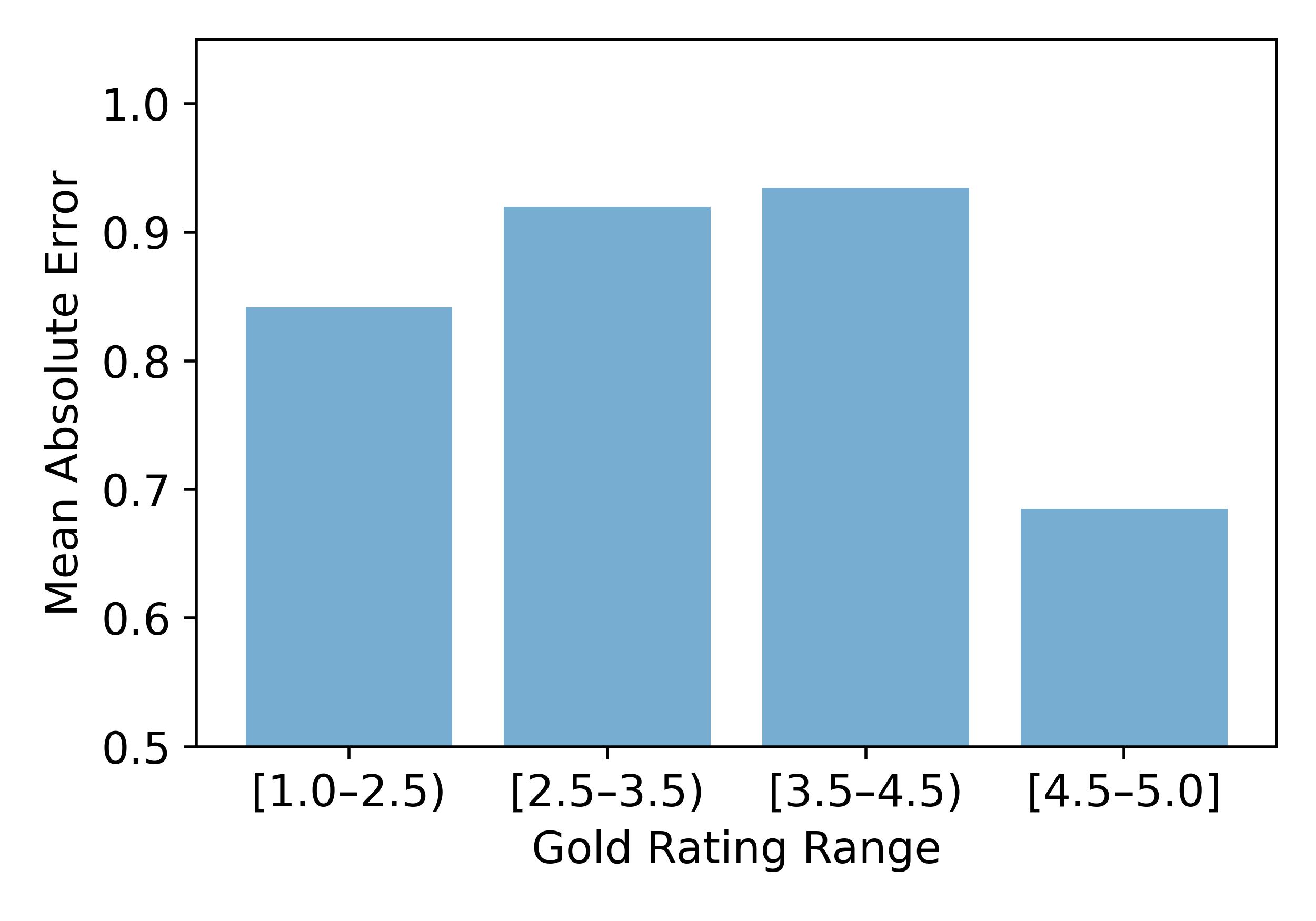}

  (a) MAE by gold rating range
\end{minipage}%
\hfill
\begin{minipage}[t]{0.49\linewidth}
  \centering
  \includegraphics[width=\linewidth]{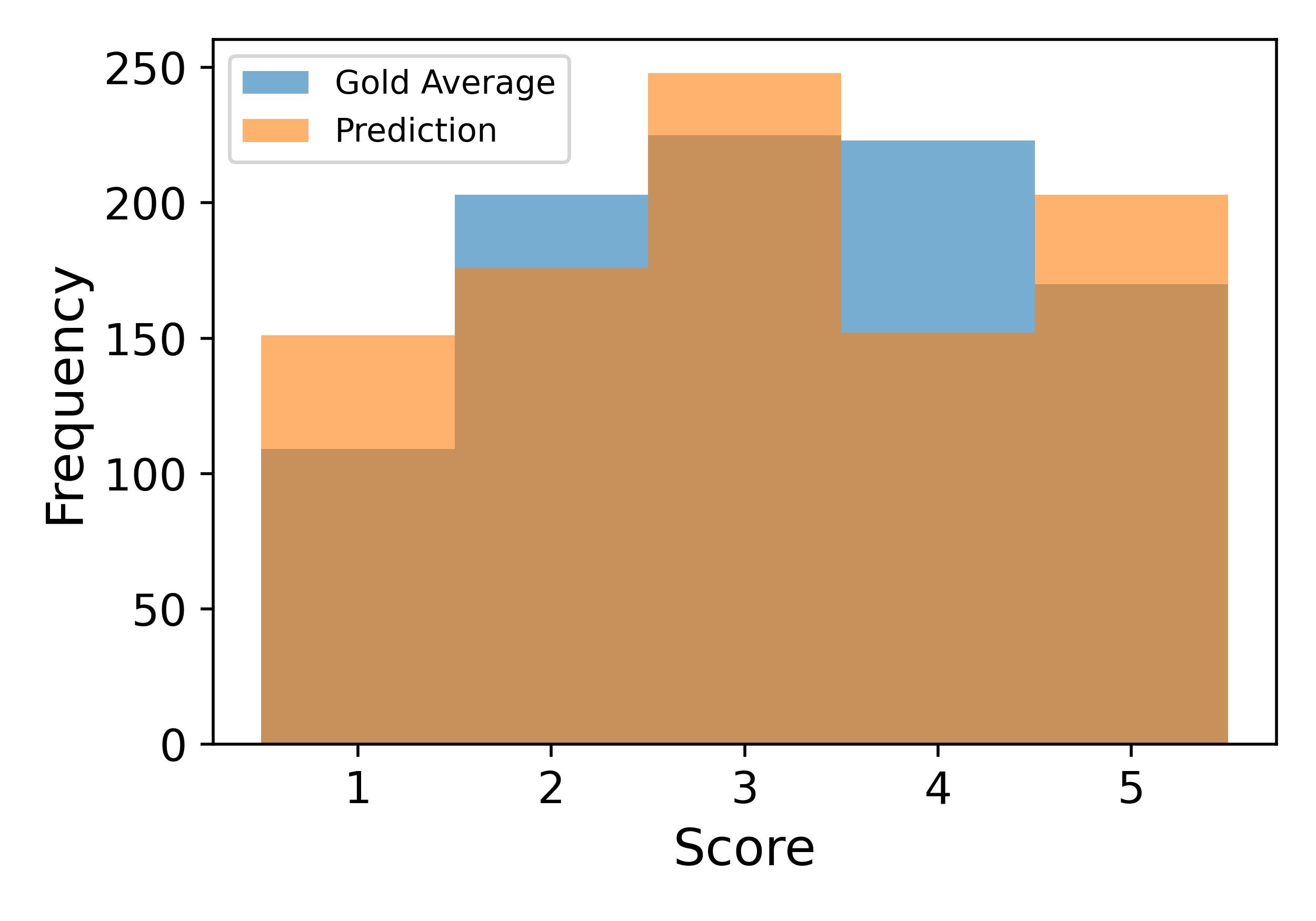}

  (b) Gold vs.\ predicted score distributions
\end{minipage}
\captionof{figure}{Error analysis of GPT-5.2 (P1) on the test set.}
\label{fig:error_analysis}

\vspace{1mm}

\begin{minipage}[t]{0.49\linewidth}
  \centering
  \includegraphics[width=\linewidth]{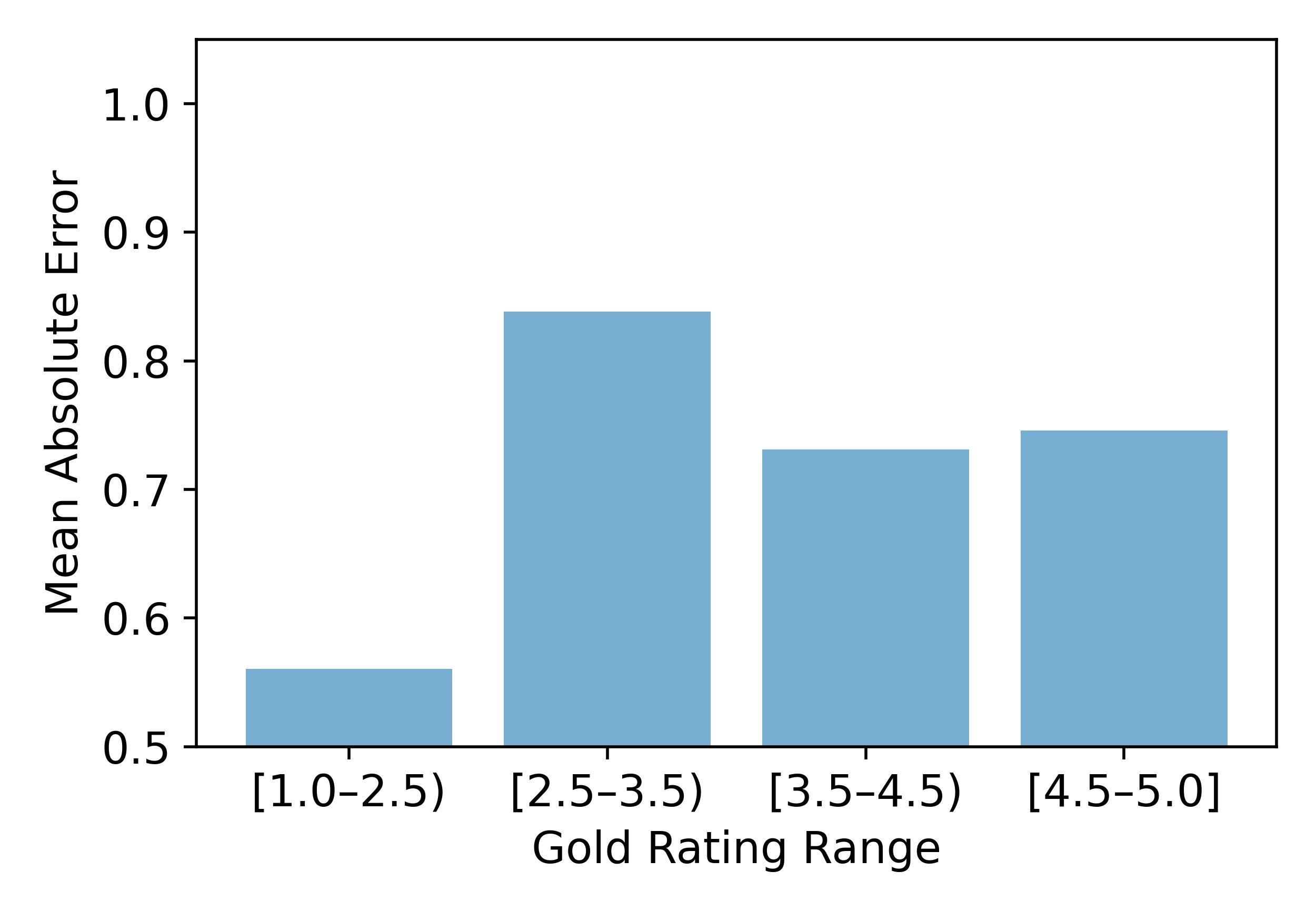}

  (a) MAE by gold rating range
\end{minipage}%
\hfill
\begin{minipage}[t]{0.49\linewidth}
  \centering
  \includegraphics[width=\linewidth]{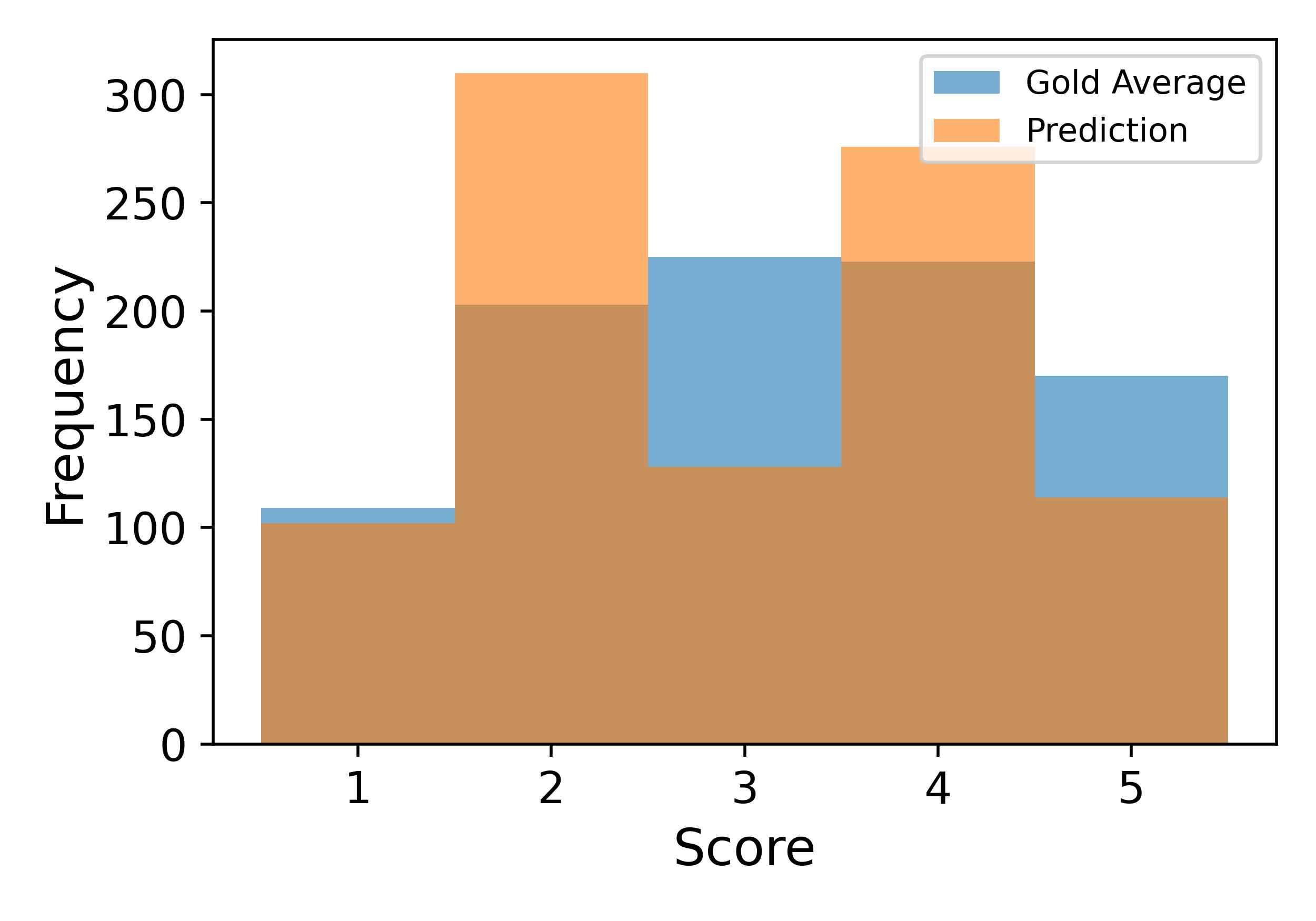}

  (b) Gold vs.\ predicted score distributions
\end{minipage}
\captionof{figure}{Error analysis of GPT-4o (P2) on the test set.}
\label{fig:error_analysis_gpt4o}
\end{minipage}
\end{center}
\end{samepage}

\paragraph{Mid-range ratings are hardest to predict.} In Figure~\ref{fig:error_analysis}, the left panel shows MAE by gold rating range. The highest MAE occurs for ratings in $[3.5, 4.5)$ (MAE = 0.934, $n = 223$), while extreme ratings near 1 or 5 are easiest (MAE $\approx$ 0.69). Extreme cases involve clear confirmation or contradiction by the ending, whereas mid-range ratings require nuanced judgment about partial evidence. The right panel of Figure~\ref{fig:error_analysis} further reveals that the model's predictions cluster at integer values (1--5), while gold ratings are continuously distributed, indicating a discretization bias that limits fine-grained estimation. This stems from the prompt instruction to return a single integer. No experiments were conducted with non-integer ratings or averaging multiple stochastic API calls, as integer outputs were simpler to parse and validate under a deterministic setting. Exploring continuous-valued prompting strategies could reduce this bias.

\paragraph{Misleading precontexts cause catastrophic errors.}
Catastrophic errors typically involve homonyms where the narrative context unambiguously supports one sense, yet the model assigns high plausibility to another. For example, \textit{shelved} in a library context led to prediction~5 for ``hold back to a later time,'' while gold~=~1.4. These failures highlight the tension between the ending-priority rule and human judgment: the model conflated meanings and rated ``hold back to a later time'' as~5, ignoring that both precontext and ending support only the ``place on a shelf'' sense. Future prompting strategies should explore more flexible weighting schemes that balance component-level evidence, accounting for sense-priming effects, rather than enforcing fixed hierarchies.

\paragraph{GPT-4o (P2) error analysis.}
Figure~\ref{fig:error_analysis_gpt4o} shows the error distribution of GPT-4o (P2), which achieves an MAE of 0.702, lower than GPT-5.2 (P1) at 0.791. Gains from decision rules appear for low-plausibility ($[1.0, 2.5)$: MAE = 0.560) and mid-range ratings ($[3.5, 4.5)$: MAE = 0.731), suggesting that calibration rules help most when judgments are ambiguous. GPT-4o maintains better alignment with the continuous gold distribution, with less concentration at boundary values, indicating that structured prompting enables more calibrated estimates. This pattern suggests that explicit decomposition into precontext, target sentence, and ending mainly improves calibration rather than raw lexical matching, which is consistent with the overall advantage of prompting over embedding-based methods.

\section{Conclusion}
This paper presents a system for SemEval-2026 Task 5, comparing embedding-based, fine-tuning, and LLM prompting approaches for word sense plausibility rating. The key contribution is a structured prompting strategy with explicit decision rules, which achieves the best performance ($\rho = 0.731$, Acc.\ = 0.794 on test). The results demonstrate that: (1) embedding-based similarity features fail to capture narrative-level reasoning; (2) fine-tuned transformers with LoRA and auxiliary losses (pairwise ranking, uncertainty) improve over standard regression; and (3) structured prompting with component-wise evaluation and calibration rules outperforms both few-shot prompting and fine-tuning. Across all comparisons, the most consistent advantage comes from making the plausibility judgment more explicit: separating precontext, target sentence, and ending gives the model a simple task-specific scaffold that is more effective than relying on global similarity or larger model size alone. The error analysis also suggests that the remaining gap is less about lexical knowledge and more about calibration under mixed evidence, especially for mid-range labels and high-disagreement cases. Future work includes exploring ensemble methods that combine fine-tuned models with LLM predictors, and improving prompts to better handle conflicts.

\section*{Limitations}
Structured prompting with decision rules improves calibration, but performance varies with prompt formulation and may require adaptation across models or settings. Evaluation is limited to English AmbiStory narratives and may not generalize to other domains, languages, or longer contexts. Paired bootstrap significance tests were conducted to assess statistical significance of test-set comparisons between systems. Predictions show concentration at integer points relative to the continuous gold ratings, which may limit output granularity and fine-grained calibration. The present study also evaluates only a small set of prompt variants, so some of the observed gains may depend on wording choices that were not explored systematically.

\section*{Acknowledgments}
Thanks to the organizers for providing the dataset, evaluation infrastructure, and coordinating the task evaluation process. 

\bibliography{task5-ref} 

\appendix
\section{Prompt}
\label{app:prompts}

\paragraph{P1: Few-Shot Prompting.}

The following shows the system message, an example, and the user prompt template. Four additional examples (ratings 2--5) follow the same format, selected from training samples with zero annotator standard deviation, ensuring that each in-context demonstration presents a consensus rating with no annotator disagreement.

\begin{tcolorbox}[title=System Prompt, colback=gray!5, colframe=gray!60, coltitle=black, fonttitle=\bfseries, left=2pt, right=2pt, top=2pt, bottom=2pt, boxrule=0.4pt]
You are evaluating whether a proposed meaning of a homonym is supported by its narrative context.

Input format:\\
- Homonym: The ambiguous word\\
- Meaning: The proposed interpretation\\
- Precontext: Background narrative\\
- Sentence: The sentence containing the homonym\\
- Ending: The conclusion (may be none)

Rating scale:\\
1 = Completely implausible. The meaning clearly conflicts with the narrative.\\
2 = Mostly implausible. Weak or contradictory support.\\
3 = Moderately plausible. Possible but ambiguous.\\
4 = Very plausible. Strong and consistent support.\\
5 = Highly plausible. Clearly intended and strongly confirmed.

The ending is the most important factor for disambiguation.\\
Return only a single integer (1--5). No explanation.
\end{tcolorbox}

\begin{tcolorbox}[title=Few-Shot Example (1 of 5), colback=gray!5, colframe=gray!60, coltitle=black, fonttitle=\bfseries, left=2pt, right=2pt, top=2pt, bottom=2pt, boxrule=0.4pt]
Homonym: drive $|$ Meaning: hitting a golf ball off of a tee with a driver\\
Precontext: Lisa had always been competitive. Every weekend, she dedicated herself to her passion. She believed that her relentless practice would pay off someday.\\
Sentence: Her drive was what ultimately got her into the top university.\\
Ending: She made that long trip to show the course coordinators her dedication to going to that university, and they said that was one of the reasons why they accepted her.\\
Rating: 1
\end{tcolorbox}

\begin{tcolorbox}[title=User Prompt, colback=gray!5, colframe=gray!60, coltitle=black, fonttitle=\bfseries, left=2pt, right=2pt, top=2pt, bottom=2pt, boxrule=0.4pt]
Homonym: \{homonym\}\\
Meaning: \{judged\_meaning\}\\
Precontext: \{precontext\}\\
Sentence: \{sentence\}\\
Ending: \{ending\}\\
Rating:
\end{tcolorbox}

\paragraph{P2: Structured Prompting with Decision Rules.}

\begin{tcolorbox}[title=System Prompt (P2), colback=gray!5, colframe=gray!60, coltitle=black, fonttitle=\bfseries, breakable, left=2pt, right=2pt, top=2pt, bottom=2pt, boxrule=0.4pt]
You are an impartial evaluator assessing whether a proposed meaning of a word is supported by the provided narrative context. Base your judgment only on the text given.

Word: \{homonym\} \quad\\ 
Proposed meaning: \{judged\_meaning\}

Narrative context\\
- Beginning (precontext): \{precontext\}\\
- Sentence containing the word: \{sentence\}\\
- Ending (conclusion): \{ending\}

Task\\
Rate the plausibility that the word \{homonym\} is used with the proposed meaning \{judged\_meaning\} in this narrative.

Evaluation criteria\\
1.\ Precontext: Does the setup make this meaning likely or expected?\\
2.\ Target sentence: Does the local usage support this meaning?\\
3.\ Ending: Does the conclusion reinforce or confirm this meaning? This is the strongest source of evidence.

Decision rules\\
- If the ending clearly contradicts the proposed meaning, the rating must be 1 or 2.\\
- If evidence is mixed or unclear, choose the lower plausible rating.\\
- A rating of 5 requires explicit confirmation in the ending and no contradictions elsewhere.

Rating scale\\
1 Completely implausible: Clear contradiction.\\
2 Mostly implausible: Weak or conflicting evidence.\\
3 Moderately plausible: Possible but ambiguous.\\
4 Very plausible: Strong and consistent support.\\
5 Highly plausible: Clearly intended and explicitly confirmed.

Output format\\
Return only a single integer from 1 to 5. Do not include explanations, comments, or any extra text.

\end{tcolorbox}

\end{document}